\documentclass[10pt,twocolumn,letterpaper]{article}

\usepackage{wacv}              

\usepackage{graphicx}
\usepackage{amsmath}
\usepackage{amssymb}
\usepackage{booktabs}

\usepackage[pagebackref,breaklinks,colorlinks]{hyperref}

\usepackage[capitalize]{cleveref}
\crefname{section}{Sec.}{Secs.}
\Crefname{section}{Section}{Sections}
\Crefname{table}{Table}{Tables}
\crefname{table}{Tab.}{Tabs.}

\def\wacvPaperID{6} 
\def\confName{WACV}
\def\confYear{2026}

\begin{document}

\title{Learning Domain Agnostic Latent Embeddings of 3D Faces for Zero-shot Animal Expression Transfer}

\author{
Yue Wang$^{1}$, Lawrence Amadi$^{2}$, Xiang Gao$^{1}$, Yazheng Chen$^{1}$, Yuanpeng Liu$^{1}$, Ning Lu$^{2}$, and Xianfeng David Gu$^{1}$\\
$^{1}$Stony Brook University\\
$^{2}$Futurewei Technologies\\
\tt\small \{wang139,gao2,yazchen,yuanpliu,gu\}@cs.stonybrook.edu\\
\tt\small \{lamadi,nlu\}@futurewei.com
}

\maketitle
\begin{figure*}[htbp]
    \centering
    \includegraphics[width=0.9\textwidth]{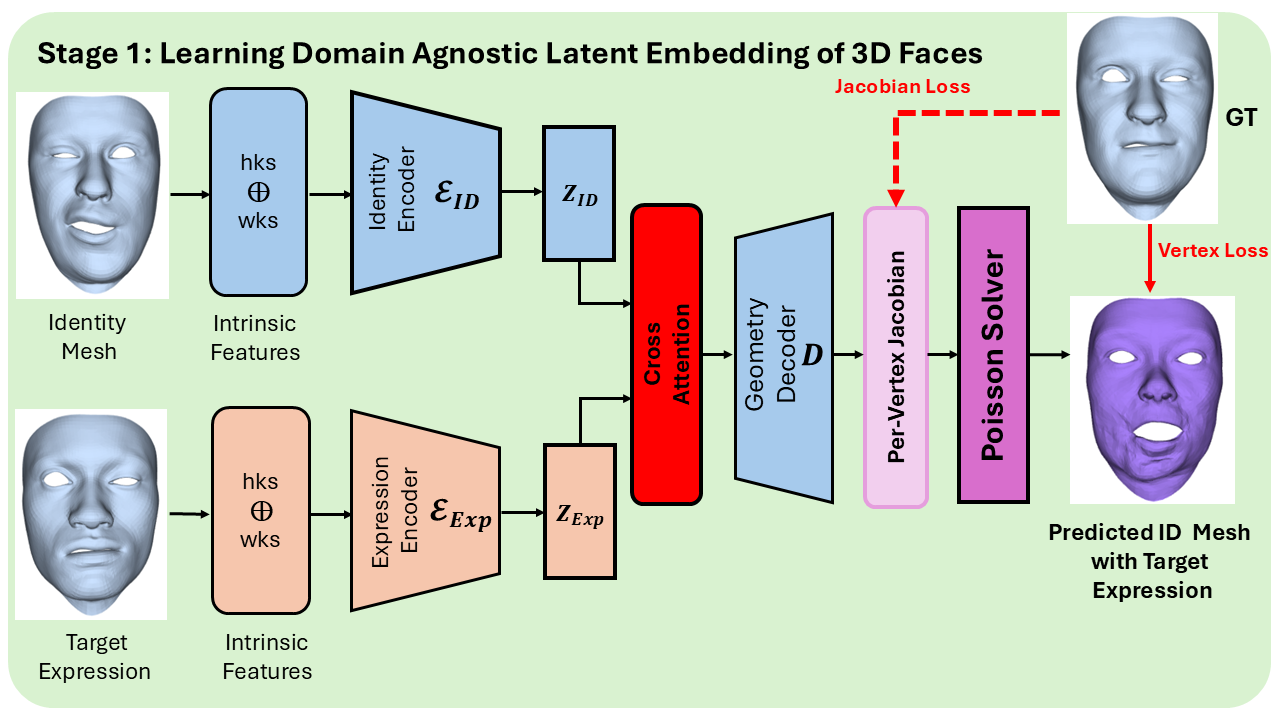} 
    \includegraphics[width=0.9\textwidth]{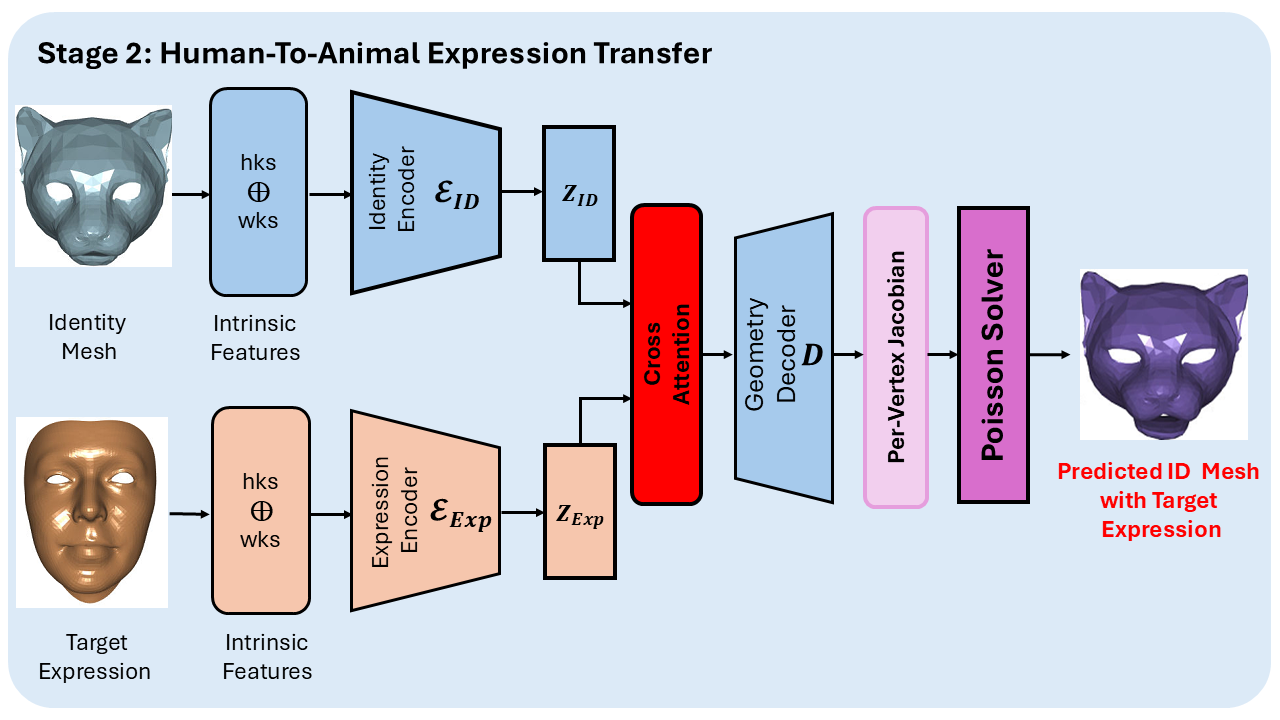} 
    \caption{
Overview of the proposed framework.
\textbf{Top: Training.} Per-vertex intrinsic features (HKS/WKS) extracted from identity and expression meshes are processed by DiffusionNet encoders, fused via cross-attention, and decoded into a per-vertex Jacobian field that defines local deformations. Training is supervised using human face triplets with both vertex position and Jacobian losses.
\textbf{Bottom: Inference.} At test time, the same model transfers expressions from a human source to an unseen animal identity in a zero-shot manner, without requiring animal expression supervision or mesh pre-alignment.
}

    \label{fig:fullwidth}
\end{figure*}
\begin{abstract}
We present a zero-shot framework for transferring human facial expressions to 3D animal face meshes. Our method combines intrinsic geometric descriptors (HKS/WKS) with a mesh-agnostic latent embedding that disentangles facial identity and expression. The ID latent space captures species-independent facial structure, while the expression latent space encodes deformation patterns that generalize across humans and animals. Trained only with human expression pairs, the model learns the embeddings, decoupling, and recoupling of cross-identity expressions, enabling expression transfer without requiring animal expression data. To enforce geometric consistency, we employ Jacobian loss together with vertex-position and Laplacian losses. Experiments show that our approach achieves plausible cross-species expression transfer, effectively narrowing the geometric gap between human and animal facial shapes.
 
\end{abstract}

\section{Introduction}
\label{sec:intro}

The rise of immersive technologies such as VR, AR, and the metaverse has driven increasing demand for highly expressive digital avatars. While existing facial animation systems predominantly focus on humans, there is growing interest in animating non-human characters and creatures. Achieving this goal requires transferring human expressions—captured via cameras or headsets—onto arbitrary non-human meshes. This task is particularly challenging due to the scarcity of paired expression datasets for animals or fictional characters, as well as the substantial morphological differences across species.

Learning-based expression transfer methods have achieved impressive results within the human domain, where aligned datasets and blendshape models are readily available. However, these approaches struggle to generalize beyond humans. Classical geometry-based techniques~\cite{gao2025inverserenderinghighgenussurface, gao2025neuralgeometryimagebasedrepresentations, gao2025GraphCut, wang2025ottalkanimating3dtalking}, including non-rigid ICP and functional maps, focus on aligning shapes but are insufficient to model expressive deformations across species. The core challenge lies in designing a representation and learning strategy that captures facial expressions in a manner that is robust to differences in species, mesh topology, and appearance.

We address this problem with a zero-shot cross-species expression transfer framework. Our model is trained solely on human expression data but generalizes to animals and other non-human identities without requiring any paired supervision. The key insight is to leverage intrinsic shape descriptors, such as heat kernel signatures (HKS) and wave kernel signatures (WKS), which encode geometry-aware features that are invariant to non-rigid deformations and mesh discretization. These descriptors are processed by a DiffusionNet~\cite{DN} encoder, which produces latent representations for both a human expression source and an animal identity target. An attention module fuses these latent features, and a neural Jacobian field predicts per-vertex deformation matrices, enforcing smooth and coherent transfers. Combining Jacobian regularization with vertex-level supervision enables stable and semantically meaningful expression transfer across species.

Our framework opens new possibilities for VR, gaming, and animation, enabling users to animate non-human characters with their own expressions in real time. Unlike prior systems that rely on hand-crafted rigs or species-specific training, our method generalizes across diverse identities, bridging the gap between human motion capture and expressive non-human avatars.

\section{Related Work}

\subsection{Human-to-Human Expression Transfer}
Early work in facial expression transfer focused on human-to-human animation using blendshapes and deformation models. Classical methods typically rely on dense correspondences or multilinear models to transfer expressions between identities~\cite{Vlasic2005FaceTransfer,Cao2014FaceWarehouse}. More recent learning-based approaches leverage neural networks to predict deformation fields or reconstruct detailed 3D geometry from images~\cite{Richardson2016Learning,Thies2016Face2Face}. While effective for human faces, these methods are limited to shared topology, often require neutral input identities, and cannot generalize across species.

\subsection{Cross-Domain and Zero-Shot Expression Transfer}
To handle cross-domain scenarios, such as transferring human expressions to avatars or synthetic characters, embedding-based latent space approaches have been proposed. Zhang et al.~\cite{zhangTVCG} use a variational autoencoder to construct separate latent spaces for human and avatar expressions, then learn a mapping between them. GAN-based methods also enable expression transfer from a single image~\cite{Siarohin2019FirstOrder,Pumarola2018GANimation}, but typically require paired examples or extensive supervision. Similarly, 3DMM-based approaches~\cite{3dmm0, 3dmm1, 3dmm2, Booth20183DMM} are effective within the human domain but struggle with zero-shot generalization to unseen identities or species. These constraints limit applicability to unfamiliar domains, such as animals.

\subsection{Mesh-Agnostic and Intrinsic Deformation Methods}
Recent works explore mesh-agnostic deformation, aiming to model geometric transformations across shapes with varying topology and discretization. DiffusionNet~\cite{DN} processes intrinsic surface features in a discretization-agnostic manner, while Neural Jacobian Fields (NJF)~\cite{NJF} learn continuous deformation fields transferable across meshes without shared topology.  

Neural Face Rigging (NFR)~\cite{NFR} builds on these ideas to transfer facial expressions across human identities. It uses intrinsic surface representations together with NJF but requires pre-aligned meshes and neutral identity inputs, limiting flexibility. LeGO~\cite{LeGO}, built on NFR, enables one-shot human face stylization via fine-tuning on a 3D Morphable Model (3DMM). While it allows identity-level stylization across diverse styles and remains compatible with blendshape pipelines, it focuses on static, human-only modifications rather than dynamic expression transfer.  

We integrate components of these approaches, and our method performs expression-driven deformation, supports expressive input identities, does not require pre-alignment, and generalizes zero-shot across species without paired examples or animal-specific training data.

\subsection{Cross-Species Deformation}
Cross-species deformation and animation remain challenging due to significant geometric, semantic, and topological differences between species. Early work by Sumner and Popovi\'c~\cite{3ddeformation} demonstrated that deformation transfer can propagate motion and pose across meshes with compatible structure, establishing a foundation for cross-domain animation. Subsequent approaches extended these ideas to animals by relying on hand-crafted correspondences or parametric templates to handle inter-species variability~\cite{zuffi2017lions, zuffi20183dsmal}.

In parallel, there has been progress in building statistical morphable models for animal faces. For example, CAFM~\cite{CAFM} constructs a 3D Morphable Model for cat-like animals, enabling automatic fitting of animal faces from images and providing a shared parametric space for shape variation. The cat face dataset used in our work is derived from this model. While CAFM focuses on parametric reconstruction rather than animation, it provides useful geometric priors for cross-species facial modeling.


\section{Method}

Our objective is to learn a representation that enables transferring facial expressions from a source mesh to a target mesh with a different identity, while remaining agnostic to mesh discretization and domain. The model is trained on human facial expressions and evaluated in a zero-shot manner on animal meshes. In addition to the expression transfer task, we analyze the learned latent spaces to better understand how identity and expression information are encoded and disentangled.

\subsection{Expression Transfer Framework}

Given a source mesh representing a facial expression and a target mesh representing an identity, our method predicts a deformation that transfers the given expression onto the target identity. To facilitate generalization across meshes with different resolutions and topologies, we operate on intrinsic geometric descriptors rather than raw vertex coordinates.

\paragraph{Intrinsic Mesh Features.}
For each mesh, we compute intrinsic spectral descriptors based on the Laplace--Beltrami operator. Specifically, we extract the Heat Kernel Signature (HKS) and Wave Kernel Signature (WKS), which are invariant to isometric deformations and robust to mesh discretization. Let $\{\lambda_i, \phi_i\}_{i=1}^K$ denote the eigenvalues and eigenfunctions of the Laplace--Beltrami operator. The HKS at vertex $v$ and time scale $t$ is defined as
\begin{equation}
\mathrm{HKS}(v,t) = \sum_{i=1}^{K} e^{-\lambda_i t} \phi_i(v)^2,
\end{equation}
capturing local diffusion behavior at multiple spatial scales. The WKS emphasizes oscillatory behavior in the spectral domain and is defined as
\begin{equation}
\mathrm{WKS}(v,e) = \sum_{i=1}^{K} g(e,\lambda_i)\phi_i(v)^2,
\end{equation}
where $g(\cdot)$ is a Gaussian window centered at logarithmic energy $e$. We use $16$ logarithmically spaced time scales for HKS and $16$ energy bands for WKS, yielding a $32$-dimensional intrinsic feature vector per vertex.

\paragraph{DiffusionNet Feature Encoding.}
Given a triangular surface mesh with $N$ vertices, the DiffusionNet encoder maps intrinsic input descriptors (e.g., HKS and WKS) to vertex-wise latent features. Each DiffusionNet layer combines diffusion-based feature propagation with nonlinear channel mixing, yielding embeddings that are intrinsically defined on the surface. Formally, let $\mathbf{X} \in \mathbb{R}^{N \times C_{\mathrm{in}}}$ denote the input descriptors and $\Delta$ the discrete Laplace--Beltrami operator. The encoder produces a latent representation $\mathbf{F} \in \mathbb{R}^{N \times C}$ through a sequence of $L$ layers, where the $l$-th layer computes
\begin{equation}
\mathbf{F}^{(l+1)} = \phi^{(l)}\!\left(\mathbf{F}^{(l)}, \exp(-t^{(l)} \Delta)\mathbf{F}^{(l)}\right),
\end{equation}
with $\mathbf{F}^{(0)} = \mathbf{X}$. Here, $t^{(l)} \in \mathbb{R}^{C}$ denotes a vector of learnable diffusion time parameters applied channel-wise, $\phi^{(l)}$ is a pointwise multilayer perceptron, and $C$ denotes the dimensionality of the latent feature space shared across all layers. This construction yields vertex-wise embeddings that encode spatially coherent geometric and deformation-related information while remaining equivariant to mesh discretization.

In the expression transfer task, the model operates directly on the vertex-wise DiffusionNet latent space without collapsing it into a single global vector. It allows expression information from a source mesh to be integrated with identity information from a target mesh at the vertex level.

\paragraph{Attention-Based Feature Fusion.}
Building on the vertex-wise latent representations produced by DiffusionNet, we design a feature fusion mechanism that transfers expression information from a source mesh to a target identity mesh while preserving spatial resolution.

Let $\mathbf{F}_s \in \mathbb{R}^{N_s \times C}$ and $\mathbf{F}_t \in \mathbb{R}^{N_t \times C}$ denote the DiffusionNet features of the source and target meshes, respectively. To incorporate global expression context without collapsing spatial information, we augment each vertex feature by concatenating it with a broadcasted global mean:
\begin{equation}
\tilde{\mathbf{F}} = \left[ \mathbf{F} \;\middle\|\; \mathbf{1}_N \left( \frac{1}{N} \sum_{i=1}^{N} \mathbf{F}_i \right)^\top \right].
\end{equation}
We then apply an attention mechanism to fuse source and target features. Given queries $\mathbf{Q}$, keys $\mathbf{K}$, and values $\mathbf{V}$ obtained from linear projections of $\tilde{\mathbf{F}}_s$ and $\tilde{\mathbf{F}}_t$, the attention output is computed as
\begin{equation}
\mathrm{Attn}(\mathbf{Q},\mathbf{K},\mathbf{V}) = \mathrm{softmax}\!\left(\frac{\mathbf{Q}\mathbf{K}^\top}{\sqrt{d}}\right)\mathbf{V},
\end{equation}
where $d$ is the feature dimensionality. This formulation enables each vertex of the identity mesh to attend to expression-relevant features of the expression mesh while maintaining vertex-wise resolution, without assuming identical vertex counts or predefined correspondences between the two meshes.

\paragraph{Jacobian-Based Deformation Prediction.}
Rather than directly predicting vertex displacements, the network outputs a per-vertex Jacobian field $\mathbf{J}_v \in \mathbb{R}^{3 \times 3}$, representing local linear deformation. Given a reference position $\mathbf{x}_v$, the deformed position is obtained by integrating the Jacobian field over the mesh, ensuring locally coherent deformations. This formulation enables the network to model complex non-rigid transformations while maintaining smoothness and physical plausibility.

\paragraph{Training Strategy and Loss Functions.}
During training, the network is exposed only to human facial meshes. Each training sample consists of a triplet: a source mesh with a given expression, a target mesh with an arbitrary initial expression, and the ground-truth mesh exhibiting the transferred expression. At inference time, the trained model is applied zero-shot to animal meshes.

We supervise the network using losses inspired by Neural Jacobian Fields (NJF). The vertex reconstruction loss penalizes positional error:
\begin{equation}
\mathcal{L}_{\mathrm{vert}} = \frac{1}{N}\sum_{v=1}^{N} \|\hat{\mathbf{x}}_v - \mathbf{x}_v\|_2^2,
\end{equation}
while the Jacobian loss enforces consistency between predicted and ground-truth local deformations:
\begin{equation}
\mathcal{L}_{\mathrm{jac}} = \frac{1}{N}\sum_{v=1}^{N} \|\hat{\mathbf{J}}_v - \mathbf{J}_v\|_F^2.
\end{equation}
The final objective is a weighted sum:
\begin{equation}
\mathcal{L} = 10\,\mathcal{L}_{\mathrm{vert}} + \mathcal{L}_{\mathrm{jac}}.
\end{equation}
This combination encourages accurate vertex reconstruction while preserving locally consistent deformation behavior.

\subsection{Latent Space Analysis and Visualization}

While the expression transfer model operates on vertex-wise features, analyzing the learned representations at the mesh level provides insight into how identity and expression are encoded. To this end, we construct global latent descriptors from the vertex-wise encoder outputs and use them solely for analysis and visualization.

Given per-vertex latent features $\mathbf{F}$, we compute a pooled mesh-level descriptor using permutation-invariant pooling:
\begin{equation}
\mathbf{z}_{\mathrm{pool}} = \mathrm{Pool}(\mathbf{F}),
\end{equation}
where $\mathrm{Pool}(\cdot)$ is implemented by concatenating mean and max pooling across vertices using \texttt{torch.cat}. These pooled descriptors are not used for expression transfer or training losses, but only for studying the structure of the latent spaces (e.g., via t-SNE).

\paragraph{Dataset and Pretraining Setup.}
The latent analysis dataset consists of 50 identities, each associated with 100 expressions. Among these, 50 expressions are shared across all identities, while the remaining 50 expressions are randomly sampled per identity. Two pretraining tasks are considered to analyze identity and expression representations separately.

\begin{figure}[t]
    \centering
    \includegraphics[width=\linewidth]{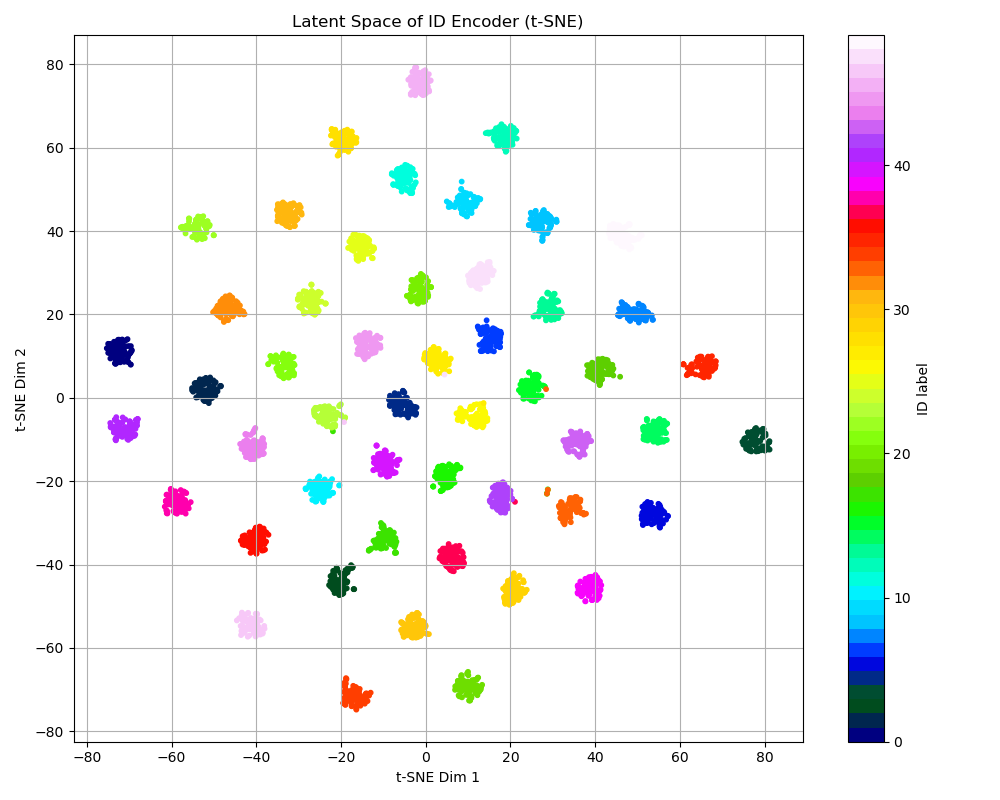}
    \caption{t-SNE visualization of the identity latent space. Points are color-coded by identity, showing clear clustering across different expressions.}
    \label{fig:id_tsne}
\end{figure}

\begin{figure}
    \centering
    \includegraphics[width=\linewidth]{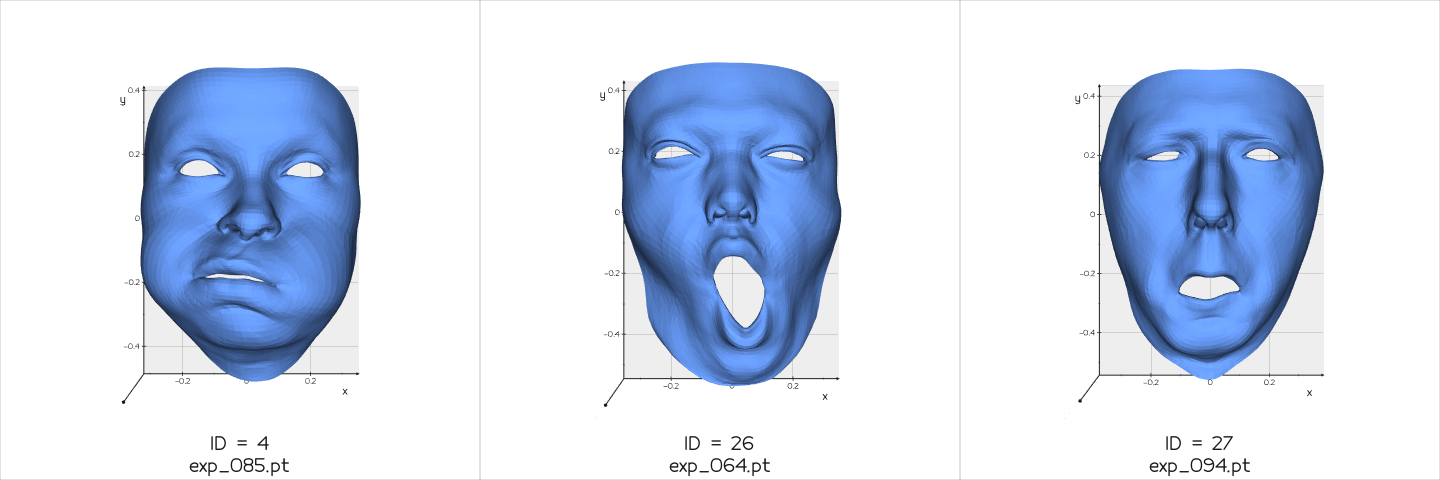}
    \includegraphics[width=\linewidth]{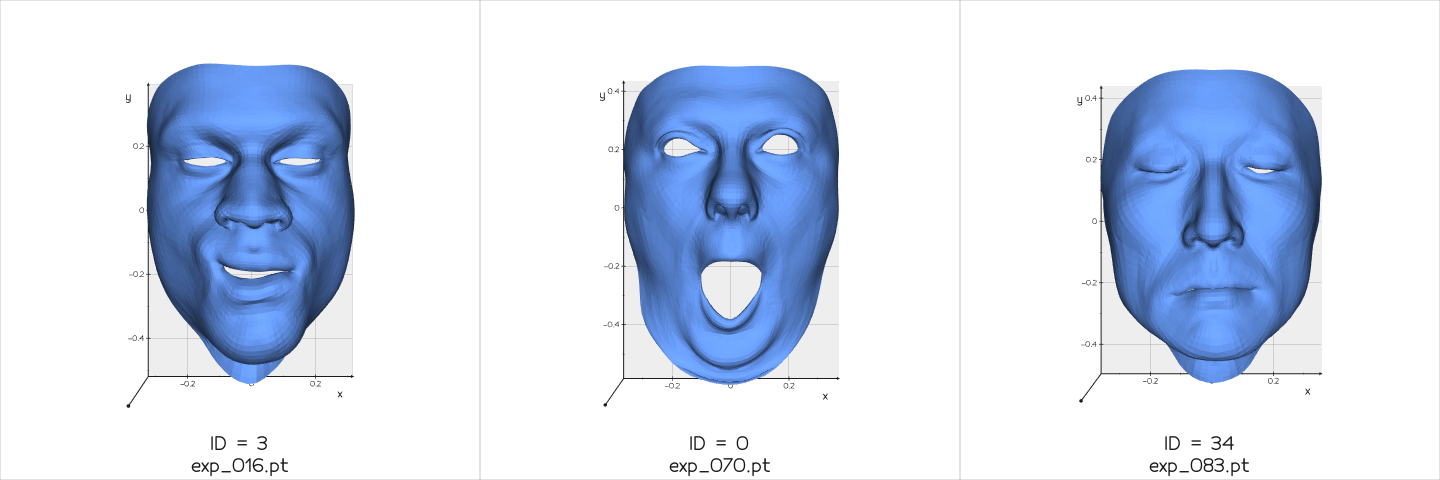}
    \caption{Top: Closest three identity pairs in the learned ID latent space. Bottom: Farthest three identity pairs, illustrating separation between distinct identities.}
    \label{fig:id_neighbors}
\end{figure}

\paragraph{Identity Latent Space.}
For identity classification, the network is trained on all 100 expressions per identity, encouraging the latent representation to cluster by identity while remaining invariant to expression. The ID latent space visualization thus includes all expressions from the 50 identities. Figure~\ref{fig:id_tsne} shows a two-dimensional t-SNE projection of the pooled latent features, where the samples are color-coded by identity class. To further illustrate the structure of the space, Figure~\ref{fig:id_neighbors} visualizes the closest and farthest identity triplets based on Euclidean distance in the pooled latent space.

\begin{figure}[t]
    \centering
    \includegraphics[width=\linewidth]{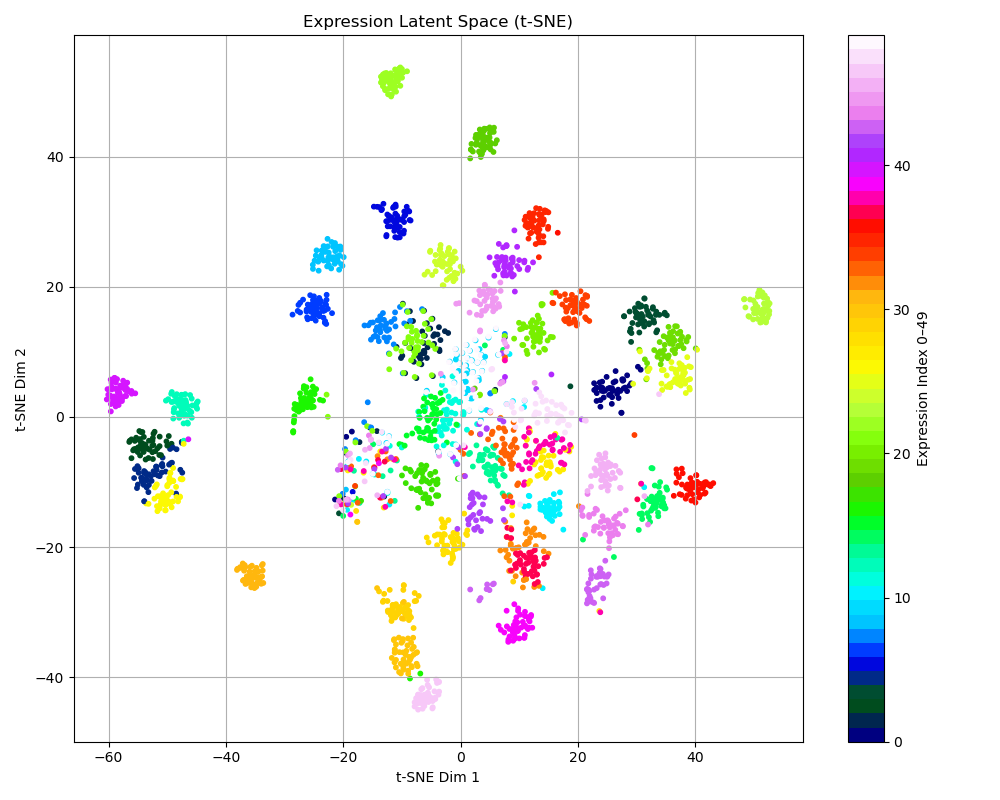}
    \caption{t-SNE visualization of the expression latent space, color-coded by shared expression category across identities.}
    \label{fig:expr_tsne}
\end{figure}

\begin{figure}
    \centering
    \includegraphics[width=\linewidth]{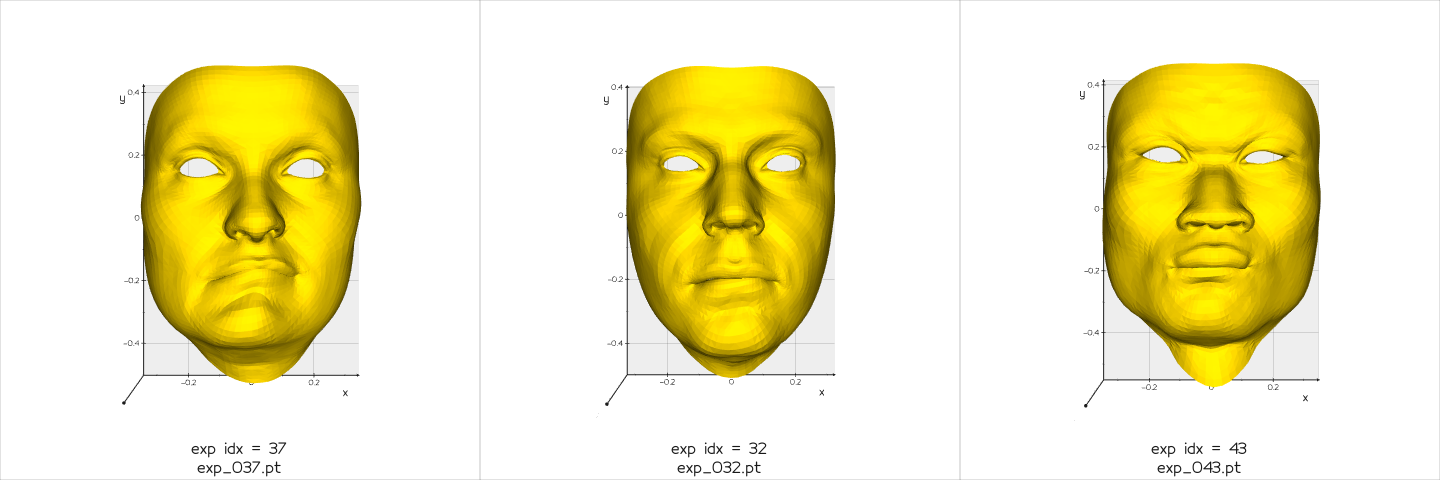}
    \includegraphics[width=\linewidth]{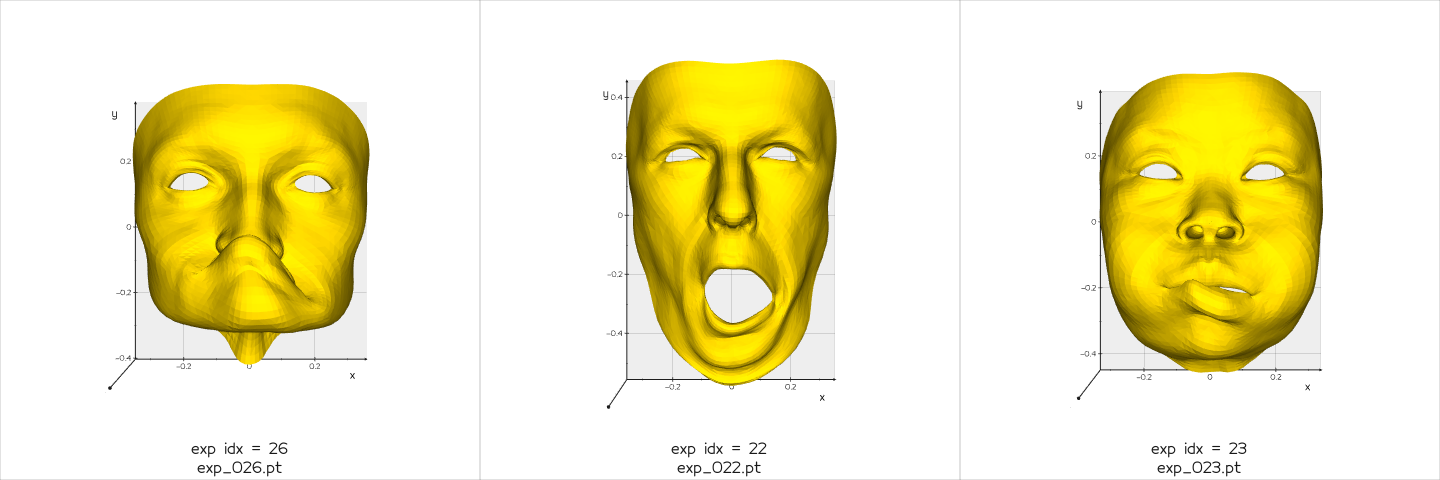}
    \caption{Top: Closest three expression pairs in the expression latent space. Bottom: Farthest three expression pairs, highlighting semantic separation between distinct facial deformations.}
    \label{fig:expr_neighbors}
\end{figure}

\paragraph{Expression Latent Space.}
For expression regression, the network is trained on the 50 random expressions per identity (expressions 50--99) to predict ICT expression parameters. To visualize expression similarity independently of identity, the remaining 50 shared expressions (expressions 0--49) are treated as expression categories during latent analysis. Figure~\ref{fig:expr_neighbors} shows the closest and farthest expression triplets in the learned expression latent space, while Figure~\ref{fig:expr_tsne} presents a t-SNE projection color-coded by expression category.

\section{Experiments}
\label{sec:experiments}

We evaluate our method on both in-domain human facial expression transfer and zero-shot cross-species transfer to animal meshes. Our experiments are designed to answer the following questions: (i) does the proposed model successfully transfer expressions between human identities, and (ii) can the same model generalize zero-shot to non-human facial geometries.

\paragraph{Training Data and Setup.}
All models are trained on a single NVIDIA RTX 5090 GPU. Training data are generated using the ICT Face Model~\cite{ICTFaceModel}, from which we synthesize $1000$ triplets of meshes of the form $(\mathcal{M}_{\text{id}}, \mathcal{M}_{\text{exp}}, \mathcal{M}_{\text{gt}})$. Each triplet consists of an identity mesh with an arbitrary initial expression, an expression-providing mesh, and the corresponding ground-truth mesh exhibiting the transferred expression. All training meshes are human faces; no animal meshes are seen during training.

\begin{figure}
    \centering
    \setlength{\tabcolsep}{2pt}
    \begin{tabular}{ccc}
        \subfloat[\\Prediction of Exp. A]{%
            \includegraphics[width=0.3\columnwidth]{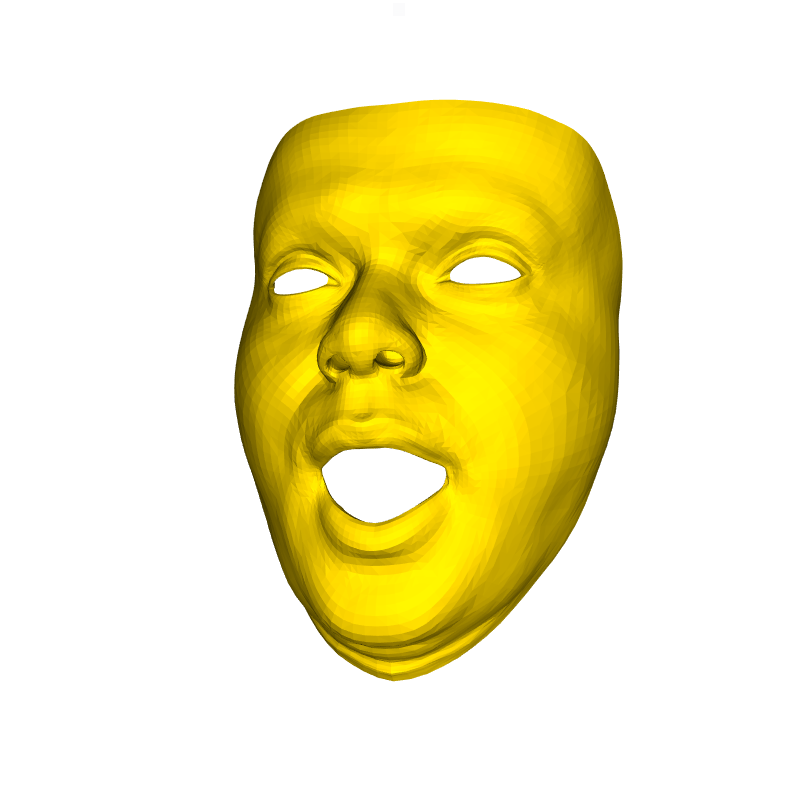}
        } &
        \subfloat[\\Prediction of Exp. B]{%
            \includegraphics[width=0.3\columnwidth]{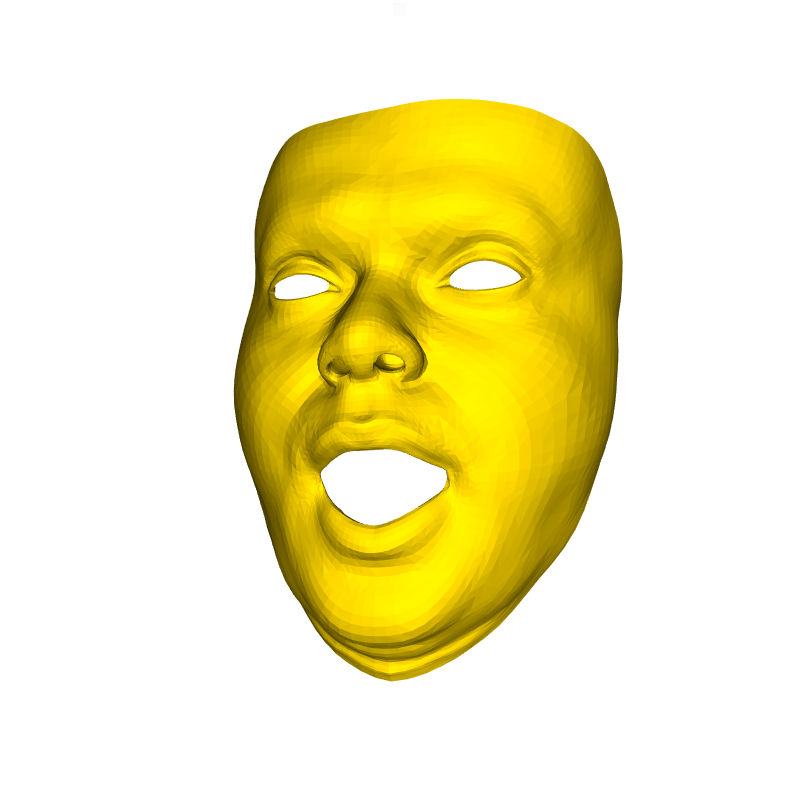}
        } &
        \subfloat[\\Prediction of Exp. C]{%
            \includegraphics[width=0.3\columnwidth]{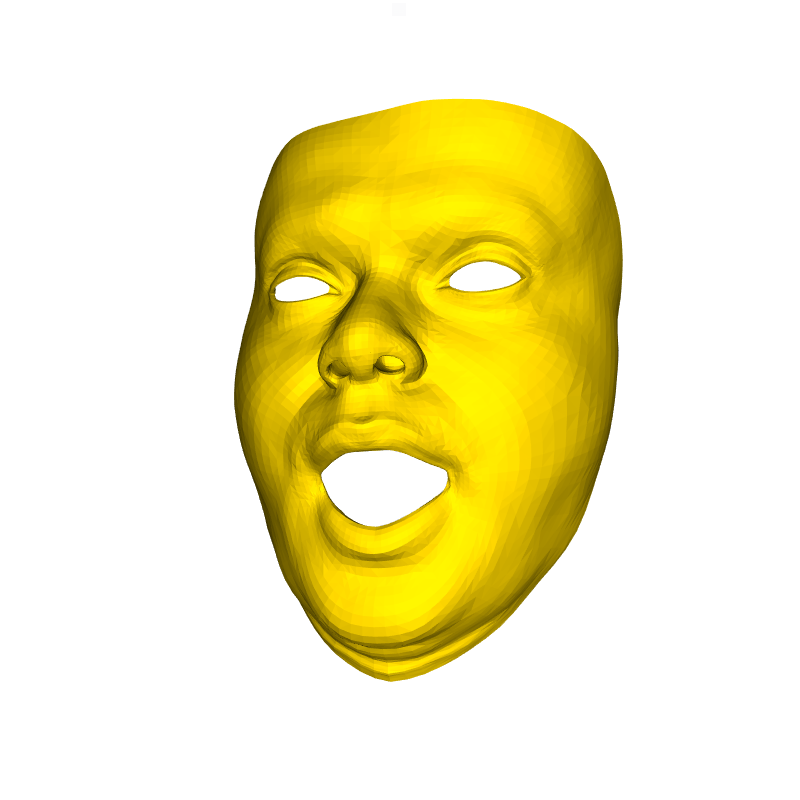}
        }
    \end{tabular}
    \caption{Predicted results for three different target expressions given the same open-mouth identity as input by NFR. The model preserves the expressive identity while adapting to varying expression sources.}
    \label{fig:nfr_comparison}
\end{figure}
\paragraph{Human-to-Human Expression Transfer.}
We first evaluate our method on human-to-human expression transfer to establish an in-domain baseline. Given a human identity mesh and a human expression mesh, the network predicts a deformation that transfers the expression onto the identity geometry. Qualitative visualizations demonstrate that the proposed approach reproduces fine-grained facial deformations, including mouth opening, eyebrow motion, and cheek deformation.
\begin{figure}[t]
    \centering
    \setlength{\tabcolsep}{2pt}

    \begin{tabular}{ccc}
        &
        \subfloat{\includegraphics[width=0.3\columnwidth]{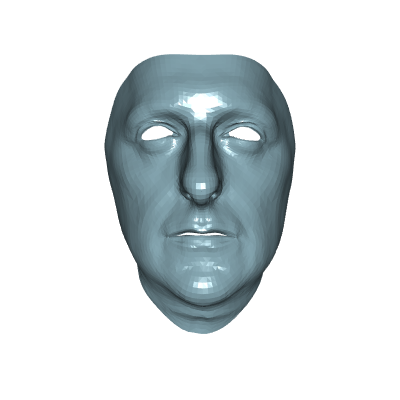}} &
        \subfloat{\includegraphics[width=0.3\columnwidth]{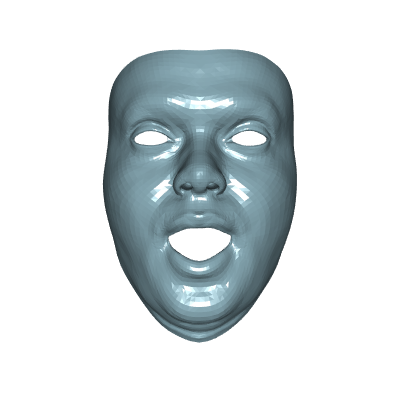}} \\

        \subfloat{\includegraphics[width=0.3\columnwidth]{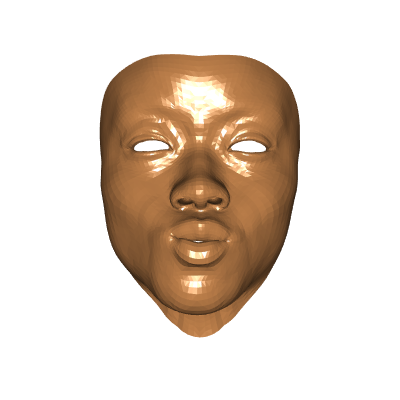}} &
        \subfloat{\includegraphics[width=0.3\columnwidth]{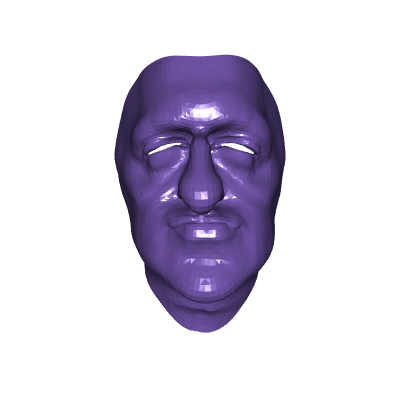}} &
        \subfloat{\includegraphics[width=0.3\columnwidth]{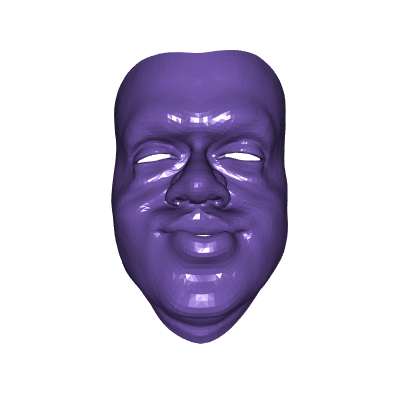}} \\

        \subfloat{\includegraphics[width=0.3\columnwidth]{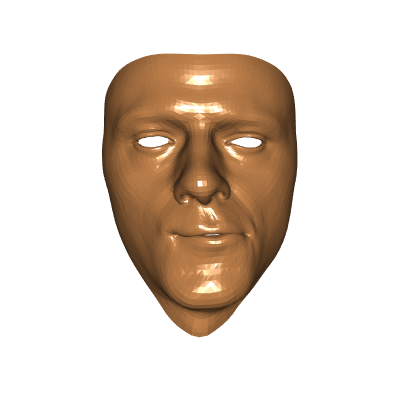}} &
        \subfloat{\includegraphics[width=0.3\columnwidth]{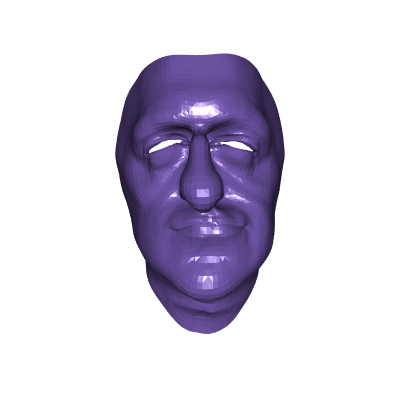}} &
        \subfloat{\includegraphics[width=0.3\columnwidth]{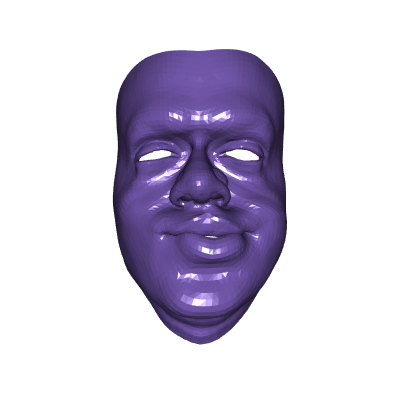}} \\

        \subfloat{\includegraphics[width=0.3\columnwidth]{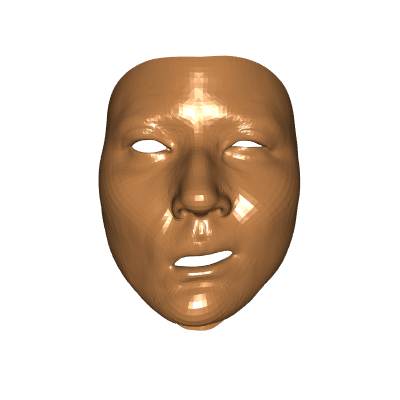}} &
        \subfloat{\includegraphics[width=0.3\columnwidth]{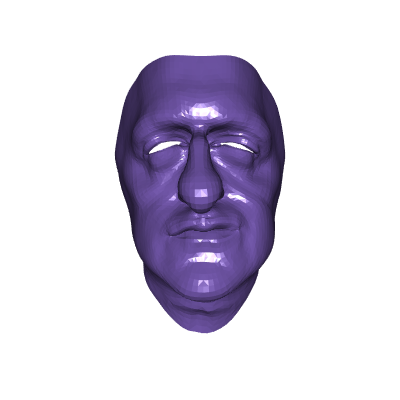}} &
        \subfloat{\includegraphics[width=0.3\columnwidth]{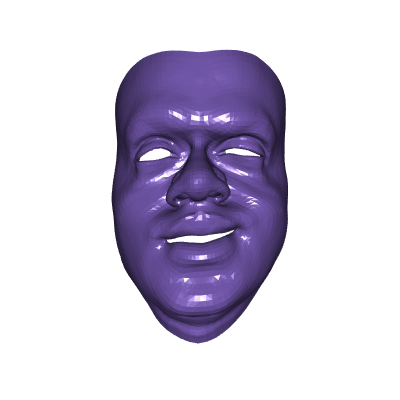}} \\

        \subfloat{\includegraphics[width=0.3\columnwidth]{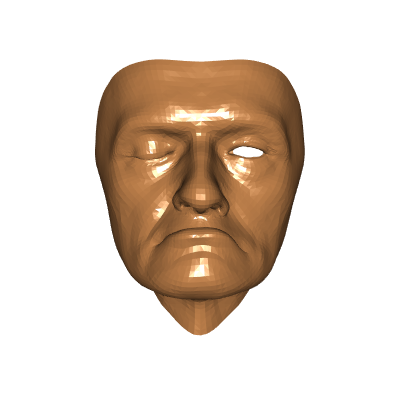}} &
        \subfloat{\includegraphics[width=0.3\columnwidth]{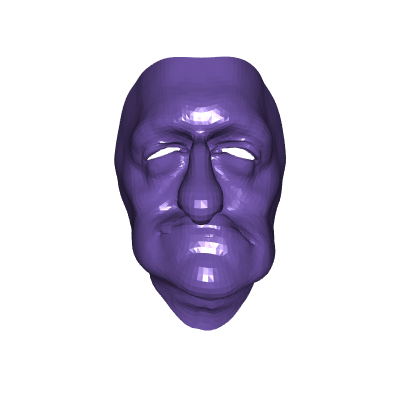}} &
        \subfloat{\includegraphics[width=0.3\columnwidth]{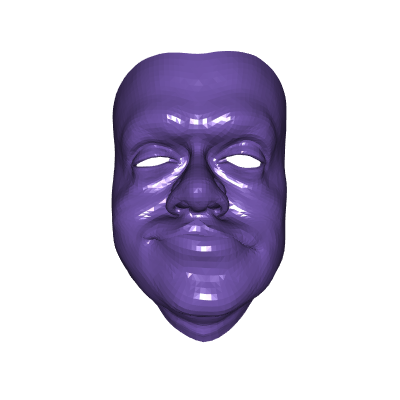}} \\

        \subfloat{\includegraphics[width=0.3\columnwidth]{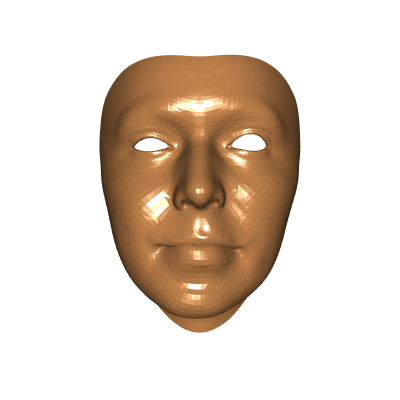}} &
        \subfloat{\includegraphics[width=0.3\columnwidth]{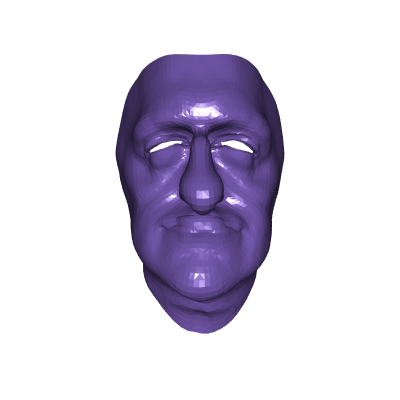}} &
        \subfloat{\includegraphics[width=0.3\columnwidth]{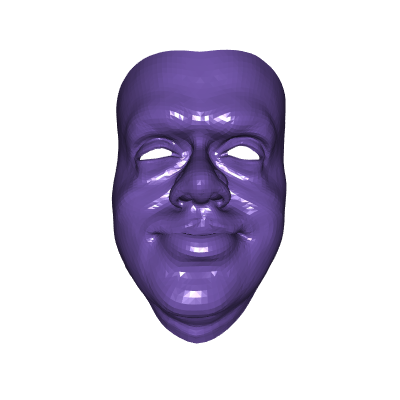}} \\

        \subfloat{\includegraphics[width=0.3\columnwidth]{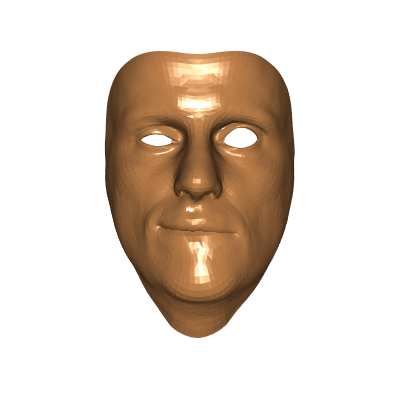}} &
        \subfloat{\includegraphics[width=0.3\columnwidth]{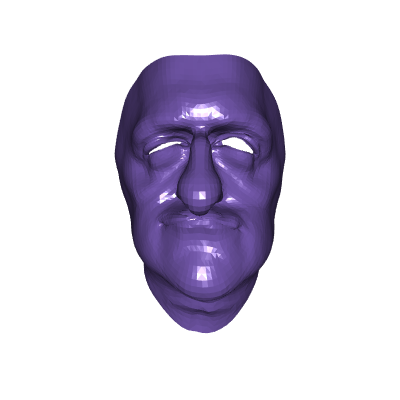}} &
        \subfloat{\includegraphics[width=0.3\columnwidth]{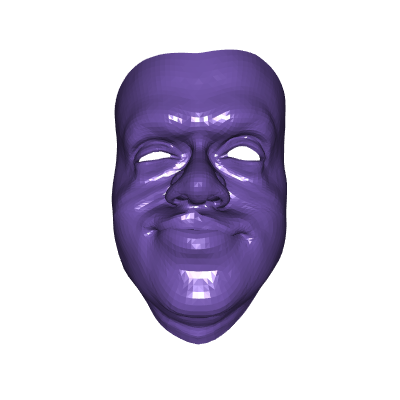}} \\
    \end{tabular}

    \caption{Qualitative results of expression editing. The first column shows target expressions; the second and third columns show results generated from neutral and open-mouth inputs, respectively.}
    \label{fig:human_qual}
\end{figure}

\begin{figure*}[t]
    \centering
    \setlength{\tabcolsep}{2pt}

    \begin{tabular}{cccccc} 
     \textbf{Target Expression}& \textbf{ID1} & \textbf{ID2} & \textbf{ID3}\\
        & 
        \subfloat{\includegraphics[width=0.2\textwidth]{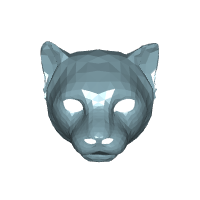}} &
        \subfloat{\includegraphics[width=0.2\textwidth]{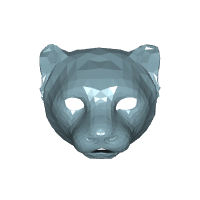}} &
        \subfloat{\includegraphics[width=0.2\textwidth]{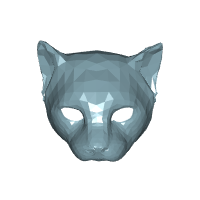}} \\  
        \subfloat{\includegraphics[width=0.2\textwidth]{figs/test/humantable/kiss.png}} &
        \subfloat{\includegraphics[width=0.2\textwidth]{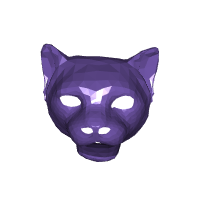}} &
        \subfloat{\includegraphics[width=0.2\textwidth]{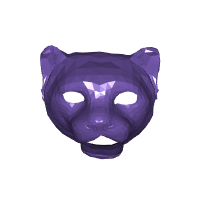}} &
        \subfloat{\includegraphics[width=0.2\textwidth]{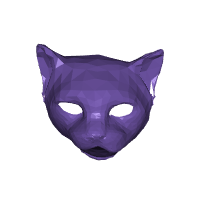}} \\

        \subfloat{\includegraphics[width=0.2\textwidth]{figs/test/humantable/mouthsideway.png}} &
        \subfloat{\includegraphics[width=0.2\textwidth]{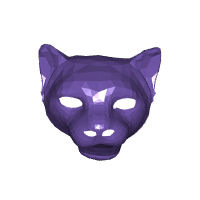}} &
        \subfloat{\includegraphics[width=0.2\textwidth]{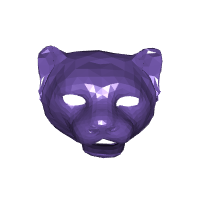}} &
        \subfloat{\includegraphics[width=0.2\textwidth]{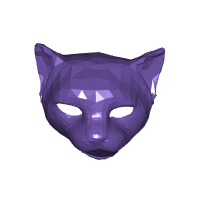}} \\



        \subfloat{\includegraphics[width=0.2\textwidth]{figs/test/humantable/smile.png}} &
        \subfloat{\includegraphics[width=0.2\textwidth]{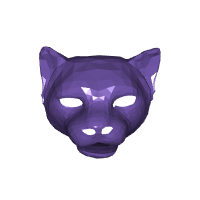}} &
        \subfloat{\includegraphics[width=0.2\textwidth]{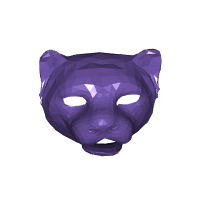}} &
        \subfloat{\includegraphics[width=0.2\textwidth]{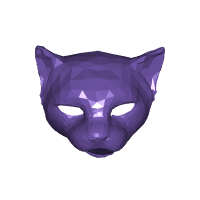}} \\

        \subfloat{\includegraphics[width=0.2\textwidth]{figs/test/humantable/1lipup1eyesquint.png}} &
        \subfloat{\includegraphics[width=0.2\textwidth]{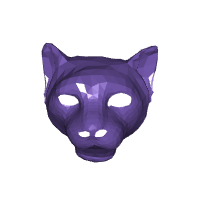}} &
        \subfloat{\includegraphics[width=0.2\textwidth]{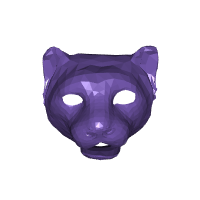}} &
        \subfloat{\includegraphics[width=0.2\textwidth]{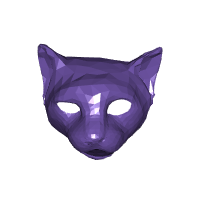}}   \\
    \end{tabular}
\captionsetup{width=0.8\textwidth} 
    \caption{\textbf{Zero-shot human-to-animal expression transfer.} The model is trained exclusively on human faces and applied directly to animal meshes at inference. Top row: different target cat identities; left column: expression inputs; middle cells: predicted meshes corresponding to each target ID and expression combination. From top to bottom, the four expression inputs are: a kiss, leftward mouth shift, smile, and an asymmetric expression with one eye enlarged, one eye reduced, and the mouth shifted to the right.}

    \label{fig:human_animal}
\end{figure*}

\paragraph{Comparison with NFR.}
We compare our method with Neural Face Rigging (NFR)~\cite{NFR}, a state-of-the-art facial deformation transfer approach that does not require shared mesh topology, making it applicable to both human and animal faces. However, NFR assumes that the input identity is provided in a neutral expression and relies on pre-aligned neutral meshes for accurate deformation transfer. As a result, identities exhibiting non-neutral expressions must be manually or algorithmically normalized prior to inference.

In contrast, our method directly operates on expressive identity meshes without requiring neutralization or pre-alignment, enabling more flexible and realistic expression transfer under diverse identity conditions. Figure~\ref{fig:nfr_comparison} illustrates this limitation by showing NFR’s outputs for an open-mouth identity (the same input used in Figure~\ref{fig:human_qual}). As expected, NFR produces nearly identical outputs across different target expressions due to its neutral-input assumption, whereas our method adapts the transferred expressions to the expressive input identity.

\paragraph{Zero-Shot Human-to-Animal Expression Transfer.}
Our main contribution lies in zero-shot expression transfer from humans to animals. At test time, we apply the trained model directly to animal meshes without any fine-tuning or retraining. Despite significant differences in geometry, topology, and vertex count, the model produces plausible expression deformations that respect the target animal’s anatomy while preserving the semantic intent of the human expression. In our experiments, ID1 (Fig.~\ref{fig:human_animal}) corresponds to the mean cat face obtained from CAFM~\cite{CAFM}, a 3D Morphable Model for Animals, which serves as a canonical target identity for expression transfer.


These results highlight the benefit of intrinsic geometric features and diffusion-based encoding, which allow the network to generalize beyond human-specific mesh structures.

\paragraph{Discussion.}
Across all experiments, the proposed framework demonstrates strong generalization despite being trained on a limited set of human-only data. The combination of intrinsic descriptors, diffusion-based latent representations, and attention-based global conditioning enables robust expression transfer without requiring correspondence, consistent topology, or species-specific training data.

\section{Conclusion and Future Work}

We have presented a zero-shot framework for transferring human facial expressions to animal 3D meshes using a mesh-agnostic latent embedding that effectively disentangles identity and expression. Trained exclusively on human expression data, our approach leverages intrinsic shape descriptors and a neural Jacobian field to achieve stable and semantically meaningful cross-species expression transfer. Experimental results demonstrate that the method produces plausible deformations across a variety of species, bridging the gap between human motion capture and expressive non-human avatars.

For future work, we plan to explore training on larger and more diverse datasets, such as MultiFace~\cite{multiface}, to further improve generalization across a wider spectrum of facial geometries and species. Incorporating additional supervision signals, including semantic or muscle-based priors, could enhance the fidelity and realism of transferred expressions. Finally, extending the framework to dynamic sequences and enabling real-time applications in VR, AR, and animation pipelines represents an exciting direction for practical deployment.



{\small
\bibliographystyle{ieee_fullname}
\bibliography{egbib}
}

\end{document}